\documentclass[letterpaper]{article} 
\usepackage[preprint]{aaai2027} 
\usepackage[hyphens]{url} 
\usepackage{graphicx} 
\usepackage{natbib} 
\usepackage{caption} 
\usepackage{algorithm}
\usepackage{algorithmic}
\usepackage{amsmath}
\usepackage{booktabs}

\title{Sequential Trajectories and Simultaneous Blending: Multi-Emotion Modeling for Instruction-Following TTS}
\author{
    Yan Zhou, Yun Hong, Yang Feng\corresponding
}
\affiliations{
    Key Laboratory of Intelligent Information Processing, Institute of Computing Technology, Chinese Academy of Sciences (ICT/CAS)\\
    State Key Laboratory of AI Safety, Institute of Computing Technology, Chinese Academy of Sciences\\
    University of Chinese Academy of Sciences, Beijing, China\\
    zhouyan23z@ict.ac.cn, fengyang@ict.ac.cn
}

\begin{document}

\maketitle

\begin{abstract}
Natural-language instructions enable flexible control of synthesized speech, yet emotional TTS systems primarily model a single utterance-level affect, leaving multi-emotion control underexplored. We study two complementary multi-emotion TTS tasks: emotion trajectory, which spans several ordered affective stages, and emotion blending, in which multiple emotions coexist throughout an utterance. These tasks expose a supervision mismatch: supervised fine-tuning (SFT) does not explicitly evaluate emotion features, while single-emotion rewards provide neither structure-aware feedback for trajectory completion nor pair-aware feedback for blending. We introduce \textbf{HybridEmo}, a post-training framework that initializes both tasks with SFT and then aligns the speech-token policy through Group Relative Policy Optimization using a sample-aware hybrid reward. For trajectory samples, segment-aligned consistency combines average and weakest-stage evidence to preserve the correctness and completeness of prescribed stages. For blending samples, a GMM-based reward combines frame-level support from the union of target-emotion anchors in an offline emotion space with an utterance-level weaker-target margin. Both branches share an ASR reward and are routed within a unified policy. On MultiEmo-Test, HybridEmo significantly improves trajectory correctness and blending intensity, without a noticeable degradation in speaker similarity. Human evaluation prefers HybridEmo to CosyVoice 3 and EmoVoice-0.5B, with nearly balanced preferences against Qwen3-TTS.\footnote{Code is available at \url{https://github.com/ictnlp/HybridEmo}.}
\end{abstract}

\section{Introduction}
\label{sec:introduction}

\begin{figure*}[t]
    \centering
    \includegraphics[width=0.96\textwidth]{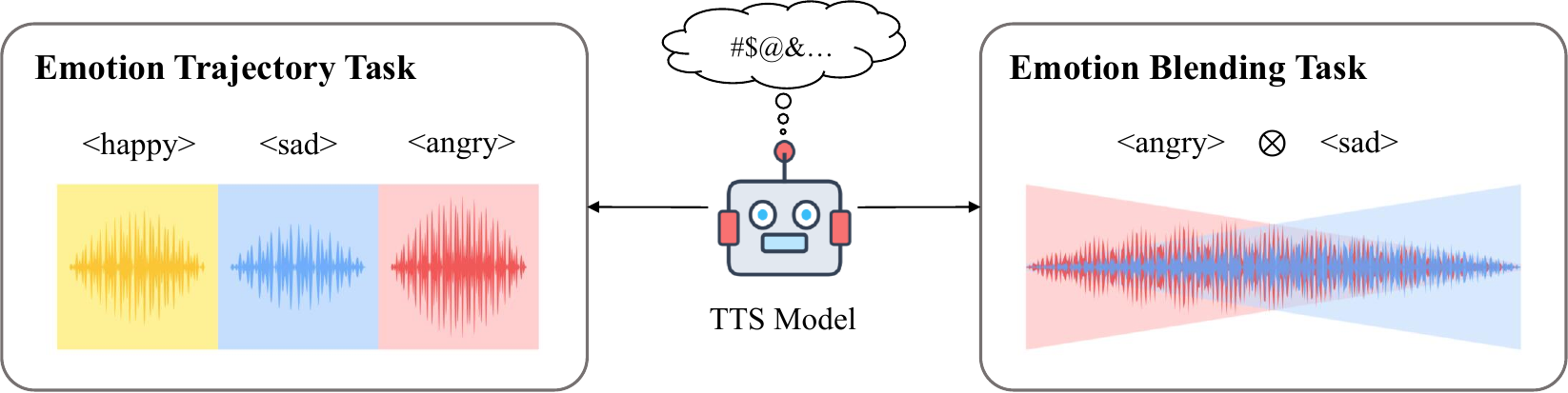}
    \caption{Multi-emotion control through sequential emotion trajectories and concurrent emotion blending.}
    \label{fig:multi_emotion_tasks}
\end{figure*}

Natural-language instructions enable flexible control of synthesized speech, but current emotional text-to-speech (TTS) systems largely model a single utterance-level emotion~\cite{du2025cosyvoice3,yang2025emovoice,chen2026flexivoice}. However, real-world utterances may contain multi-emotion patterns which can be categorized into two forms: sequential evolution and simultaneous blending. On these grounds, we study \emph{multi-emotion control in TTS}: given the text, an emotion instruction and a reference voice, the task is to generate intelligible speech following the requested emotion pattern while retaining the reference timbre. The two forms of the multi-emotion control are defined as follows (illustrated in Figure~\ref{fig:multi_emotion_tasks}): \emph{Emotion Trajectory} arranges multiple emotions sequentially within an utterance, whereas \emph{Emotion Blending} expresses multiple emotions concurrently. Both forms require modeling relationships among emotions beyond a single utterance-level label. Therefore, training a TTS model with multi-emotion control requires endowing it with the ability to capture the precise emotional patterns within speech.

However, conventional token-level supervised fine-tuning (SFT) alone is inadequate for multi-emotion control. Multi-emotion TTS requires both stable speech generation and structured emotional-pattern realization. Learning both from large-scale triplets of input text, emotion instructions, and multi-emotion target speech is costly, while cross-entropy fits target speech-token sequences without explicitly assessing the prescribed sequential or concurrent pattern. SFT therefore provides a necessary initialization but insufficient direct supervision for precise multi-emotion control.

Reinforcement learning (RL) offers a complementary solution by explicitly evaluating emotional patterns in generated speech. Starting from an SFT-initialized model, RL can optimize the speech-token policy by scoring its generated waveforms, without requiring a paired target waveform for each condition. This makes RL a better match for multi-emotion instructions that admit diverse valid realizations, while allowing emotion optimization to be combined with content-preservation objectives. However, existing speech RL objectives cannot adequately evaluate the requested multi-emotion patterns. Recent methods optimize global properties such as intelligibility and speaker similarity~\cite{sun2025f5rtts,liu2025grpotts}, or apply GRPO to flexible style control~\cite{chen2026flexivoice}, but do not explicitly evaluate relationships among multiple emotions. Moreover, emotion recognizers typically provide utterance-level scores for individual classes~\cite{ma2023emotion2vec}. For trajectories, such scores cannot localize emotion changes or detect missing and misordered emotions; for blending, they cannot measure whether multiple targets coexist. The key challenge is therefore to construct task-matched rewards for sequential and concurrent emotional patterns.

To meet this challenge, we propose \textbf{HybridEmo}, a two-stage post-training framework that initializes multi-emotion generation through SFT and then aligns the speech-token policy through Group Relative Policy Optimization (GRPO) with a sample-aware hybrid reward. For trajectories, a segment-aligned consistency reward combines mean and weakest-stage emotion evidence to assess both the overall correctness and the completion of prescribed emotional stages. For blending, a GMM-based mixture-density reward combines graded frame-level target-pair compatibility with an utterance-level weaker-target margin, encouraging both emotions to coexist while preventing single-target dominance. A sample-aware router applies the appropriate emotion reward to each sample type, while a shared ASR reward preserves linguistic content during emotion optimization.

We construct MultiEmo-Test to evaluate emotion trajectory and blending tasks. HybridEmo significantly improves trajectory correctness and blending-oriented perceptual scores without a noticeable degradation in speaker similarity. Human evaluation also favors HybridEmo over CosyVoice 3 and EmoVoice-0.5B, with a near-balanced preference compared with Qwen3-TTS.

Our main contributions are as follows:
\begin{itemize}
    \item We formulate multi-emotion control through emotion trajectories and blending, and introduce the MultiEmo-Test evaluation set.
    \item We propose HybridEmo, a two-stage post-training framework that combines supervised initialization with sample-aware GRPO, routing task-matched emotion rewards alongside shared content feedback.
\end{itemize}

\section{Background}
\label{sec:background}

\subsection{Discrete Speech-Token TTS}

Discrete speech tokens provide a common interface between speech waveforms and language-model-based generation. VALL-E~\cite{wang2023valle} casts zero-shot TTS as conditional language modeling over neural-codec codes, while SPEAR-TTS~\cite{kharitonov2023spear} factoarizes generation into text-to-semantic and semantic-to-acoustic stages to exploit audio-only data. The CosyVoice series~\cite{du2024cosyvoice2,du2025cosyvoice3} combines an autoregressive speech-token large language model (LLM) with a flow-matching acoustic generator for controllable synthesis and zero-shot voice cloning. Spark-TTS~\cite{wang2025sparktts} proposes BiCodec to encode linguistic content and speaker attributes in a decoupled single stream, while Qwen3-TTS~\cite{hu2026qwen3tts} develops complementary tokenizers for high-fidelity and streaming synthesis.

\subsection{Instruction-Based and Emotional TTS}

Natural-language prompting replaces fixed style labels with a more expressive way of control. PromptTTS~\cite{guo2022prompttts} conditions TTS on textual descriptions of style, and PromptTTS~2~\cite{leng2023prompttts2} adds a variation network to model vocal factors underspecified by text. InstructTTS~\cite{yang2023instructtts} maps free-form style prompts into a discrete acoustic latent space while disentangling style, speaker, and content. ControlSpeech~\cite{ji2025controlspeech} combines content, style, and speech prompts in a decoupled codec space for simultaneous speaker and style control. Specializing toward affective expression, EmoVoice~\cite{yang2025emovoice} uses an LLM to interpret fine-grained emotion descriptions and predicts phoneme and audio tokens in parallel for content consistency. Collectively, these methods broaden controllability from closed attribute inventories to free-form descriptions, but largely treat style or emotion as an utterance-level condition rather than explicitly modeling sequential emotion trajectories or simultaneous emotion blending.

\subsection{Reinforcement Learning for Speech Generation}

Reinforcement learning (RL) complements token-level supervision with sequence-level preference or reward signals that better reflect perceptual and task-specific speech quality. Direct Preference Optimization (DPO)~\cite{rafailov2023dpo} learns directly from preference pairs without fitting an explicit reward, and Emo-DPO~\cite{gao2024emodpo} adapts it to emotional TTS by constructing preferences that sharpen distinctions among target emotions. Group Relative Policy Optimization (GRPO)~\cite{shao2024deepseekmath} instead estimates relative advantages from groups of sampled outputs without a critic. F5R-TTS~\cite{sun2025f5rtts} applies GRPO to flow-matching TTS with intelligibility and speaker-similarity rewards, while another GRPO-based approach~\cite{liu2025grpotts} optimizes an LLM-based TTS model with ASR rewards. CosyVoice 3~\cite{du2025cosyvoice3} introduces differentiable reward optimization for speech-token RL, and FlexiVoice~\cite{chen2026flexivoice} progressively combines multimodal DPO with multi-objective and instruction-focused GRPO. These methods, however, largely formulate rewards over global features, providing limited structure-aware supervision for multi-stage emotion trajectory or blended target pairs.

\section{Methodology}
\label{sec:method}

\subsection{Task Definition}

Given text $x$, a natural-language emotion instruction $c_e$, and a timbre reference utterance $y_{\mathrm{ref}}$, a conditional TTS model $F_\theta$ generates
\begin{equation}
\hat y=F_\theta(x,c_e,y_{\mathrm{ref}}).
\label{eq:task_definition}
\end{equation}
The output $\hat y$ should preserve the linguistic content of $x$ and the timbre of $y_{\mathrm{ref}}$ while realizing the emotional pattern specified by $c_e$.

We consider two task types. An emotion trajectory condition contains an ordered sequence
\begin{equation}
\mathcal{T}=\{(x_k,e_k)\}_{k=1}^{K},
\qquad x=x_1\oplus\cdots\oplus x_K,
\label{eq:trajectory_condition}
\end{equation}
where $x_k$ is a text span and $e_k$ is its target emotion. The generated speech should express these emotions in the prescribed order. An emotion blending condition instead contains an unordered pair $\mathcal{B}=\{A,B\}$ of distinct non-neutral emotions that should coexist throughout the utterance; it has no ordered emotion-tagged spans. These structures require order-aware trajectory control and target-pair-aware blending control, respectively.

\subsection{Two-Stage Multi-Emotion Post-Training}

HybridEmo builds on CosyVoice 3\footnote{\url{https://github.com/FunAudioLLM/CosyVoice}}~\cite{du2025cosyvoice3}, which combines an autoregressive speech-token LLM with a downstream flow-matching acoustic generator. Across both SFT and RL, we optimize only the LLM and keep the speech tokenizer and acoustic generator frozen. The LLM is conditioned only on the input text and emotion instruction; the timbre reference is used only by the acoustic generator for waveform decoding.

\paragraph{Supervised multi-emotion initialization.}
We first apply supervised fine-tuning (SFT) to trajectory and blending demonstrations so that the model acquires an initial ability to generate both multi-emotion patterns. For a target waveform $y^*$, the frozen tokenizer produces target speech tokens $z^*$. We optimize the token-level cross-entropy objective
\begin{equation}
\mathcal{L}_{\mathrm{SFT}}
=-\sum_{t=1}^{|z^*|}
\log p_\theta
\left(z_t^*\mid z_{<t}^*,x,c_e\right).
\label{eq:sft_objective}
\end{equation}
This stage learns a conditional speech-token distribution from real multi-emotion speech before reward-based alignment.

\begin{figure*}[t]
    \centering
    \includegraphics[width=0.98\textwidth]{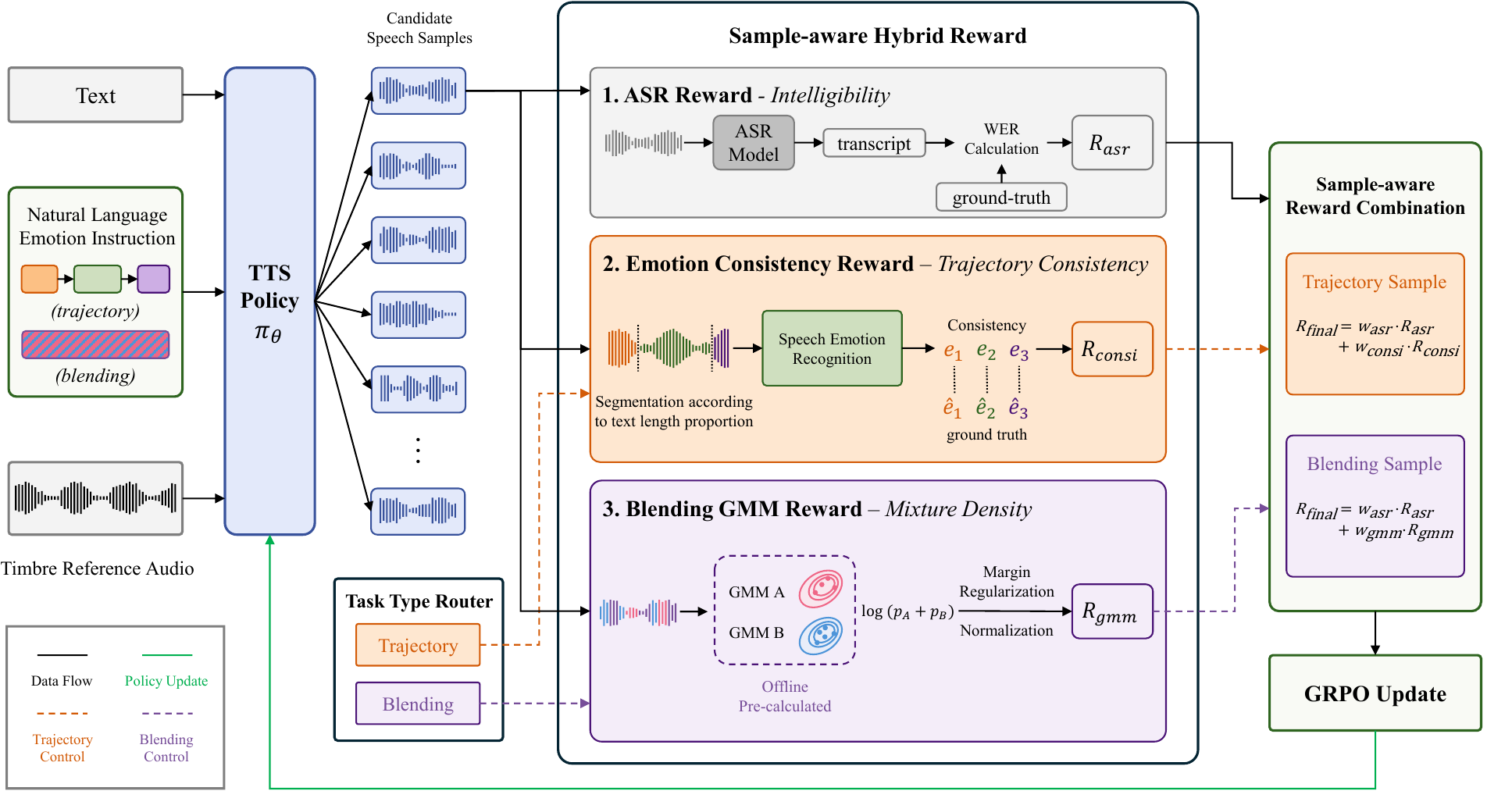}
    \caption{HybridEmo GRPO alignment. A shared ASR reward preserves linguistic content, while the task router selects either trajectory consistency or GMM-based mixture-density feedback before GRPO updates the TTS policy.}
    \label{fig:hybridemo}
\end{figure*}

\paragraph{Reinforcement learning alignment.}
Starting from the SFT checkpoint, we treat the autoregressive speech-token LLM as a policy $\pi_\theta$. A trajectory RL condition provides $(x,c_e,\mathcal{T})$, whereas a blending condition provides $(x,c_e,\mathcal{B})$; neither contains a target waveform. The policy samples a group of speech-token rollouts, which the frozen acoustic generator decodes into waveforms conditioned on $y_{\mathrm{ref}}$:
\begin{equation}
\begin{aligned}
z^{(i)}
&\sim\pi_\theta(\cdot\mid x,c_e),\\
\hat y^{(i)}
&=\mathcal{D}(z^{(i)};y_{\mathrm{ref}}),
\qquad i=1,\ldots,G,
\end{aligned}
\label{eq:generation}
\end{equation}
where the fixed acoustic generator $\mathcal{D}$ renders each rollout as speech. The sample-aware reward below scores each waveform, and GRPO~\cite{shao2024deepseekmath} computes group-relative advantages and updates the policy under a KL constraint. Figure~\ref{fig:hybridemo} illustrates this alignment stage.

\subsection{Sample-Aware Hybrid Reward}

Multi-emotion samples require different evidence according to their task structure. Let $s\in\{\mathrm{trajectory},\mathrm{blending}\}$ denote the sample type. HybridEmo shares an ASR reward $R_{\mathrm{asr}}$ across both types and routes one task-specific emotion reward:
\begin{equation}
\begin{aligned}
R(\hat y,s)
&=w_{\mathrm{asr}}R_{\mathrm{asr}}\\
&\quad+
\begin{cases}
w_{\mathrm{consi}}R_{\mathrm{consi}},
&s=\mathrm{trajectory},\\[2pt]
w_{\mathrm{gmm}}R_{\mathrm{gmm}},
&s=\mathrm{blending}.
\end{cases}
\end{aligned}
\label{eq:hybrid_reward}
\end{equation}
Here $R_{\mathrm{consi}}$ evaluates ordered trajectory completion, whereas $R_{\mathrm{gmm}}$ evaluates compatibility with a blended target pair. The inapplicable emotion component is masked, and invalid or degenerate token sequences receive zero reward.

\subsubsection{Shared ASR Reward}

Following the ASR-based content reward in CosyVoice 2~\cite{du2024cosyvoice2}, we explicitly protect intelligibility during emotion optimization. We transcribe $\hat y$ with SenseVoice\footnote{\url{https://github.com/FunAudioLLM/SenseVoice}}~\cite{an2024funaudiollm} and normalize the reference and recognized text with the same English text normalizer $\operatorname{Norm}(\cdot)$. Let $x_{\mathrm{asr}}=\operatorname{ASR}(\hat y)$, and let $\epsilon_{\mathrm{wer}}$ denote the WER between $\operatorname{Norm}(x)$ and $\operatorname{Norm}(x_{\mathrm{asr}})$. We compute
\begin{equation}
R_{\mathrm{asr}}=\operatorname{clip}_{[0,1]}\!\left(
1-\tanh(3\epsilon_{\mathrm{wer}})\right).
\label{eq:asr_reward}
\end{equation}
This bounded transformation gives high reward to content-faithful speech during emotion optimization.

\subsubsection{Trajectory-Aligned Emotion Consistency}

A trajectory condition provides the ordered emotion-tagged spans $\mathcal{T}=\{(x_k,e_k)\}_{k=1}^{K}$ but no target waveform or timestamps. Because a global emotion score cannot localize an incorrect stage or identify missing and misordered stages, we approximate span boundaries in the generated waveform from text length. Let $\ell_k=\max(|x_k|,1)$ and $D$ be the duration of $\hat y$. Stage $k$ occupies
\begin{equation}
I_k=\left[
D\frac{\sum_{j=1}^{k-1}\ell_j}{\sum_{j=1}^{K}\ell_j},
D\frac{\sum_{j=1}^{k}\ell_j}{\sum_{j=1}^{K}\ell_j}
\right).
\label{eq:trajectory_alignment}
\end{equation}
For each segment, a speech emotion recognition model, emotion2vec+ large\footnote{\url{https://huggingface.co/emotion2vec/emotion2vec_plus_large}}~\cite{ma2023emotion2vec} yields the target posterior $p_k=P(e_k\mid\hat y_k)$. Let $\bar p=K^{-1}\sum_{k=1}^{K}p_k$ and $p_{\min}=\min_k p_k$ denote the mean and weakest-stage scores. The trajectory consistency reward is
\begin{equation}
R_{\mathrm{consi}}=\beta\bar p+(1-\beta)p_{\min}.
\label{eq:consi_reward}
\end{equation}
Here $\beta\in[0,1]$ balances the mean and weakest-stage scores, preventing a high average from hiding a failed stage. For a single span, $R_{\mathrm{consi}}$ reduces to utterance-level target-emotion consistency; for multiple spans, it protects the weakest stage when evaluating trajectory completion. This timestamp-free alignment assumes that text length roughly tracks speaking duration.

\subsubsection{GMM-Based Mixture-Density Reward}

A blending condition provides the unordered target pair $\mathcal{B}=\{A,B\}$, without emotion-tagged spans or a target waveform. Because an utterance-level single-emotion posterior cannot represent this concurrent target, we construct offline frame-level GMM anchors and use their mixture-density support as graded target-pair compatibility feedback, regularized with a weaker-target margin.

\paragraph{Offline anchor construction.}
We use the base CosyVoice 3 model to synthesize single-emotion speech from instructions sampled from the SFT source corpus. We retain samples whose target-emotion confidence from emotion2vec+ large is at least $\tau$, trim boundary frames, and extract frame-level emotion features. From these labeled features, we learn an LDA projector $W$, remove outliers with Isolation Forest~\cite{liu2008isolationforest}, and fit an emotion-specific GMM in the projected space:
\begin{equation}
p_e(u)=\sum_{m=1}^{M}\omega_{e,m}
\mathcal{N}(u;\mu_{e,m},\Sigma_{e,m}).
\label{eq:gmm_anchor}
\end{equation}
Here $\omega_{e,m}\geq0$ and $\sum_{m=1}^{M}\omega_{e,m}=1$. We freeze $W$ and the per-emotion density anchors during RL. 


\paragraph{Online mixture-density scoring.}
For a blending candidate, we obtain frame features $(h_1,\ldots,h_T)$ and project them with the same frozen mapper, $u_t=Wh_t$. Given target emotions $A$ and $B$, $p_A(u_t)$ and $p_B(u_t)$ are their per-frame densities. We define an unnormalized target-pair union score in log space using $\operatorname{LSE}(a,b)=\log(\exp a+\exp b)$:
\begin{equation}
\ell_{\mathrm{mix}}(u_t)=\operatorname{LSE}\!\left(
\log p_A(u_t),\log p_B(u_t)\right).
\label{eq:mixture_term}
\end{equation}
This score is high for frames that lie in regions supported by target anchors. To discourage utterance-level dominance by one target, we compute the pair-normalized contribution of each emotion and its weaker utterance-level share:
\begin{equation}
\begin{aligned}
q_{t,e}
&=\frac{p_e(u_t)}{p_A(u_t)+p_B(u_t)},
\quad e\in\{A,B\},\\
m
&=\min_{e\in\{A,B\}}\frac{1}{T}\sum_{t=1}^{T}q_{t,e}.
\end{aligned}
\label{eq:weaker_target_share}
\end{equation}
We then regularize the utterance-level union score with a normalized weaker-target margin:
\begin{equation}
\begin{aligned}
s_{\mathrm{union}}
&=\frac{1}{T}\sum_{t=1}^{T}\ell_{\mathrm{mix}}(u_t),\\
s_{\mathrm{raw}}
&=s_{\mathrm{union}}-\alpha\max\left(0,1-\frac{m}{\rho}\right).
\end{aligned}
\label{eq:raw_gmm_reward}
\end{equation}
Here $\rho\in(0,0.5]$ sets the minimum desired weaker target-anchor contribution, and $\alpha\geq0$ controls the maximum penalty. The weaker-target margin becomes inactive once $m\geq\rho$, avoiding a strict equal-density constraint.
Because the raw score is scale-sensitive, we standardize it using a running mean and scale:
\begin{equation}
\tilde{s}=
\frac{s_{\mathrm{raw}}-\mu_{\mathrm{run}}}
{\max(\sigma_{\mathrm{run}},\varepsilon)}.
\label{eq:gmm_normalization}
\end{equation}
Here $\varepsilon$ is a small constant for numerical stability. We then map the standardized score to a bounded reward:
\begin{equation}
R_{\mathrm{gmm}}=\operatorname{sigmoid}(\tilde{s}).
\label{eq:gmm_reward}
\end{equation}
After warm-up, $\mu_{\mathrm{run}}$ and $\sigma_{\mathrm{run}}$ are updated with exponential moving averages of the raw score and its absolute deviation, respectively. The resulting reward combines graded union support with a weaker-target margin while accommodating multimodal variation within each emotion. Building on the blending distribution learned during SFT, GRPO uses this dense feedback to refine outputs toward regions supported by the requested emotion pair.

\section{Experiments}
\label{sec:experiments}

\subsection{Datasets}

Table~\ref{tab:data_statistics} summarizes the data used for the two training stages and evaluation. Our SFT corpus is constructed from the English portion of the in-the-wild Emilia dataset\footnote{\url{https://huggingface.co/datasets/amphion/Emilia-Dataset}}~\cite{he2024emilia}. We use Qwen3-Omni-30B-A3B-Captioner\footnote{\url{https://huggingface.co/Qwen/Qwen3-Omni-30B-A3B-Captioner}}~\cite{xu2025qwen3omni} to describe acoustic and paralinguistic characteristics, and then use MiniMax-M2.5\footnote{\url{https://huggingface.co/MiniMaxAI/MiniMax-M2.5}}~\cite{minimax2026m25} to screen emotional speech, assign the 7-class emotion taxonomy (\emph{angry, disgusted, fearful, happy, sad, surprised}, and \emph{neutral}), identify trajectory or blending structures, and generate natural-language instructions. Trajectory examples contain 1 to 3 labeled text spans, whereas each blending example contains exactly 2 distinct non-neutral emotions.

We construct \textbf{MultiEmo-RL} for reinforcement learning. It contains 12,000 emotion trajectory and 2,400 emotion blending text--instruction pairs generated from diverse metadata with MiniMax-M2.5. Each condition consists of a target text, a natural-language instruction, and its emotion structure, without a target waveform. Trajectory examples additionally include an emotion-tagged transcript only for constructing the alignment reward. The blending portion covers 12 unordered emotion pairs.

We further construct \textbf{MultiEmo-Test} with 720 conditions. Its trajectory portion contains 200 examples for each of the 1-, 2-, and 3-stage trajectory settings, while its blending portion contains 120 examples over the same 12 emotion pairs. Each condition is paired with a reference voice from the English Seed-TTS evaluation set\footnote{\url{https://github.com/BytedanceSpeech/seed-tts-eval}}~\cite{anastassiou2024seedtts}; the same reference is provided to every system that supports timbre conditioning. We programmatically verify that MultiEmo-RL and MultiEmo-Test contain no duplicate samples.

\begin{table}[t]
\centering
{\small
\setlength{\tabcolsep}{2.2pt}
\begin{tabular}{llrrr}
\toprule
Dataset & Statistic & Traj. & Blend. & Total \\
\midrule
SFT corpus & Utterances & 119{,}981 & 4{,}115 & 124{,}096 \\
 & Duration (h) & 372.4 & 11.2 & 383.6 \\
MultiEmo-RL & Samples & 12{,}000 & 2{,}400 & 14{,}400 \\
MultiEmo-Test & Samples & 600 & 120 & 720 \\
\bottomrule
\end{tabular}
}
\caption{Statistics of the SFT corpus, MultiEmo-RL, and MultiEmo-Test. Audio duration applies only to SFT; the latter two contain text--instruction samples.}
\label{tab:data_statistics}
\end{table}

\subsection{Implementation Details}

We initialize HybridEmo from the 0.5B CosyVoice 3 checkpoint~\cite{du2025cosyvoice3} and perform SFT for 5 epochs using Adam~\cite{kingma2015adam}, with a learning rate of $2\times10^{-6}$ and an effective global batch size of 64.

We then train the model for 1 GRPO epoch on MultiEmo-RL using verl\footnote{\url{https://github.com/verl-project/verl}}~\cite{sheng2025hybridflow}. For each input, the policy samples $n=8$ speech-token rollouts at temperature 0.6. For waveform decoding and reward evaluation, the frozen acoustic generator uses a randomly sampled LibriSpeech utterance\footnote{\url{https://www.openslr.org/12}}~\cite{panayotov2015librispeech} as the voice prompt. GRPO uses a learning rate of $10^{-6}$, a batch size of 64, and KL regularization against the SFT reference policy. Both stages run on 4 NVIDIA H800 GPUs. We set $\beta=0.7$, $w_{\mathrm{asr}}=0.5$, $w_{\mathrm{consi}}=0.5$, $w_{\mathrm{gmm}}=0.3$, $\alpha=0.01$, and $\rho=0.05$.

For offline anchor construction, we use $\tau=0.8$, trim $\delta=5\%$ from each utterance boundary, project frame features to $d=6$ dimensions with LDA, and remove $\rho_{\mathrm{out}}=8\%$ of the frames using Isolation Forest. Each emotion is modeled by a full-covariance GMM with $M=3$ components.

\subsection{Baselines}

We compare HybridEmo with the 0.5B CosyVoice 3~\cite{du2025cosyvoice3} model and representative external systems: EmoVoice~\cite{yang2025emovoice} at 0.5B and 1.5B scales, and Qwen3-TTS-12Hz-1.7B-VoiceDesign~\cite{hu2026qwen3tts}. All systems synthesize the same target texts. CosyVoice 3, HybridEmo, and EmoVoice receive the same natural-language emotion instruction and reference speech.

Because its tested interface does not jointly accept a speech prompt and a text instruction, Qwen3-TTS uses a textual emotion instruction with a designated voice. Reference speech can introduce acoustic style priors that affect emotion control~\cite{chen2026flexivoice}; thus, this setting is not strictly matched to the speech-prompted systems, and we treat Qwen3-TTS as an external capability reference.

\begin{table*}[!t]
\centering
{\small
\setlength{\tabcolsep}{2.7pt}
\begin{tabular}{lcccccccccc}
\toprule
& \multicolumn{4}{c}{Correctness $\uparrow$}
& \multicolumn{3}{c}{Naturalness $\uparrow$}
& & & \\
\cmidrule(lr){2-5}\cmidrule(lr){6-8}
Model & 1E & 2E & 3E & \makebox[3.2em][c]{\shortstack{1--3E\\Avg.}} &
\makebox[2.4em][c]{2E} & \makebox[2.4em][c]{3E} & \makebox[3.2em][c]{\shortstack{2--3E\\Avg.}} &
WER (\%) $\downarrow$ & UTMOS $\uparrow$ & SIM $\uparrow$ \\
\midrule
EmoVoice-0.5B & 3.70 & 2.94 & 3.16 & 3.27 & \makebox[2.4em][c]{2.26} & \makebox[2.4em][c]{2.57} & \makebox[2.4em][c]{2.41} & 4.29 & 3.22 & 0.56 \\
EmoVoice-1.5B & 3.74 & 2.95 & 3.13 & 3.28 & \makebox[2.4em][c]{2.29} & \makebox[2.4em][c]{2.49} & \makebox[2.4em][c]{2.39} & 7.94 & 3.12 & 0.56 \\
\midrule
Qwen3-TTS$^{\ast}$ & \textbf{4.02} & 2.99 & \textbf{3.30} & \textbf{3.44} & \makebox[2.4em][c]{\textbf{2.41}} & \makebox[2.4em][c]{\textbf{2.73}} & \makebox[2.4em][c]{\textbf{2.57}} & 2.69 & \textbf{3.28} & -- \\
\midrule
CosyVoice 3 & 3.68 & 2.95 & 3.09 & 3.24 & \makebox[2.4em][c]{2.30} & \makebox[2.4em][c]{2.49} & \makebox[2.4em][c]{2.40} & 1.91 & 3.20 & \textbf{0.69} \\
\textbf{HybridEmo} & 3.78 & \textbf{3.01} & 3.21 & 3.33 & \makebox[2.4em][c]{2.40} & \makebox[2.4em][c]{2.60} & \makebox[2.4em][c]{2.50} & \textbf{1.87} & 3.21 & 0.66 \\
\bottomrule
\end{tabular}
}
\caption{Automatic evaluation results for the trajectory family on MultiEmo-Test. 1--3E Avg. is the macro correctness over the 1-, 2-, and 3-stage settings; 2--3E Avg. is the macro naturalness over the 2- and 3-stage settings because naturalness is defined only for multi-stage trajectories. SIM stands for speaker similarity. Qwen3-TTS$^{\ast}$ uses a designated voice without a speech prompt.}
\label{tab:trajectory_main}
\end{table*}

\begin{table}[!t]
\centering
{\small
\setlength{\tabcolsep}{1.4pt}
\begin{tabular}{lccccc}
\toprule
Model & Int. $\uparrow$ & Nat. $\uparrow$ &
WER (\%) $\downarrow$ & UTMOS $\uparrow$ & SIM $\uparrow$ \\
\midrule
EmoVoice-0.5B & 3.57 & 3.13 & 1.89 & 3.13 & 0.56 \\
EmoVoice-1.5B & 3.55 & 3.13 & 4.15 & 3.06 & 0.56 \\
\midrule
Qwen3-TTS$^{\ast}$ & \textbf{3.71} & \textbf{3.29} & 0.64 & \textbf{3.39} & -- \\
\midrule
CosyVoice 3 & 3.47 & 3.06 & \textbf{0.30} & 3.18 & \textbf{0.70} \\
\textbf{HybridEmo} & \textbf{3.71} & 3.24 & 0.37 & 3.17 & 0.64 \\
\bottomrule
\end{tabular}
}
\caption{Automatic blending results on MultiEmo-Test. Int. and Nat. stand for intensity and naturalness. Qwen3-TTS$^{\ast}$ uses a designated voice without a speech prompt.}
\label{tab:blending_main}
\end{table}

\begin{figure}[!t]
    \centering
    \includegraphics[width=0.93\columnwidth]{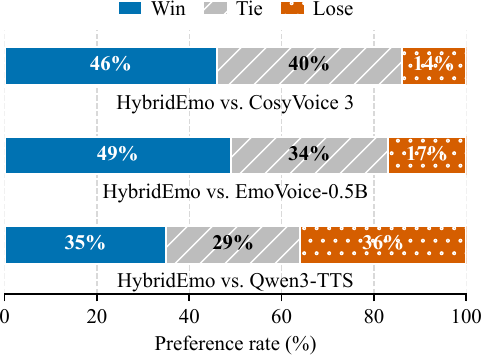}
    \caption{Pairwise preferences from human listeners on MultiEmo-Test from HybridEmo's perspective.}
    \label{fig:human_preference}
\end{figure}

\subsection{Evaluation Protocol}

We score emotion control with task-specific 1--5 rubrics using Qwen3-Omni-30B-A3B-Instruct\footnote{\url{https://huggingface.co/Qwen/Qwen3-Omni-30B-A3B-Instruct}}~\cite{xu2025qwen3omni}. For trajectories, \emph{Emotion Correctness} evaluates target expression for 1-stage samples and ordered completion for 2- and 3-stage samples, while \emph{Emotion Naturalness} evaluates within-stage expression and transition coherence for the latter two settings. We report per-setting scores with macro averages over 1--3 stages for correctness and 2--3 stages for naturalness.

For blending, \emph{Emotion Intensity} measures the overall perceptual strength of the requested two-emotion blend, while \emph{Emotion Naturalness} measures whether the two emotional qualities coexist naturally rather than appearing sequentially or collapsing to one emotion. We evaluate them jointly as task-aligned perceptual criteria.

We report corpus-level WER (\%) computed with \texttt{whisper-large-v3}\footnote{\url{https://huggingface.co/openai/whisper-large-v3}}~\cite{radford2022whisper}, applying the same English normalization to hypotheses and references. We assess acoustic quality with UTMOSv2\footnote{\url{https://huggingface.co/sarulab-speech/UTMOSv2}}~\cite{baba2024utmosv2}, which predicts a mean opinion score, and speaker similarity as the cosine similarity between ERes2Net\footnote{\url{https://modelscope.cn/models/iic/speech_eres2net_sv_en_voxceleb_16k}}~\cite{chen2023eres2net} embeddings of the reference and generated speech. Speaker similarity is unavailable for Qwen3-TTS because it uses no reference speech.

\section{Results and Analysis}
\label{sec:results}

\subsection{Automatic Evaluation Results}

HybridEmo significantly raises trajectory macro correctness from 3.24 to 3.33 over CosyVoice 3, with consistent gains at every length: 3.68 to 3.78 for 1E, 2.95 to 3.01 for 2E, and 3.09 to 3.21 for 3E. It also raises macro naturalness from 2.40 to 2.50, keeps WER below 2\% at 1.87\%, and increases UTMOS from 3.20 to 3.21, without a noticeable degradation in speaker similarity.

For blending, HybridEmo significantly raises Intensity from 3.47 to 3.71 and Naturalness from 3.06 to 3.24 over CosyVoice 3. It exceeds both EmoVoice variants on these perceptual metrics, matches Qwen3-TTS in intensity, and trails it by only 0.05 in naturalness. HybridEmo also keeps WER below 2\% at 0.37\% and UTMOS within 0.01 of CosyVoice 3, without a noticeable degradation in speaker similarity.

\subsection{Human Preference Evaluation}

Four trained listeners with synthetic-speech evaluation experience conducted pairwise comparisons for the three model pairs on MultiEmo-Test. Each listener assigned a win, tie, or loss from HybridEmo's perspective to each of 20 randomly sampled conditions per pair (70\% trajectory, 30\% blending), yielding 80 judgments per pair. Listeners considered emotion correctness, expressiveness, naturalness, and intelligibility.

Figure~\ref{fig:human_preference} shows that HybridEmo receives more wins than losses against CosyVoice 3 and EmoVoice-0.5B, with win rates exceeding loss rates by 32 percentage points in both comparisons. Against Qwen3-TTS, HybridEmo obtains 35\% wins, 29\% ties, and 36\% losses, yielding an essentially balanced preference, although the comparison is affected by differing timbre conditions. The trend is consistent with the automatic evaluation: HybridEmo surpasses CosyVoice 3 and EmoVoice and achieves a comparable preference split against Qwen3-TTS.

\subsection{Ablation Study}

We next separate the effects of SFT initialization and structure-specific reinforcement learning. Table~\ref{tab:training_analysis} uses \emph{Direct Hybrid-GRPO} for hybrid-reward GRPO initialized directly from CosyVoice 3, and \emph{Trajectory-GRPO} and \emph{Blending-GRPO} for SFT-initialized variants optimized only on the corresponding type. HybridEmo uses the same SFT initialization followed by joint sample-aware hybrid GRPO. To isolate the effect of the weaker-target margin, we also evaluate a HybridEmo variant without this term in the reward.

\begin{table}[tb]
\centering
{\small
\setlength{\tabcolsep}{3.0pt}
\begin{tabular*}{\columnwidth}{@{\extracolsep{\fill}}lccc@{}}
\toprule
\multicolumn{4}{l}{\textit{(a) Emotion trajectory}} \\
\cmidrule(lr){1-4}
Variant & Corr. $\uparrow$ & Nat. $\uparrow$ & WER (\%) $\downarrow$ \\
\midrule
CosyVoice 3 & 3.24 & 2.40 & 1.91 \\
\midrule
SFT & 3.28 & 2.47 & 2.06 \\
Direct Hybrid-GRPO & 3.30 & 2.47 & 2.08 \\
Trajectory-GRPO & \textbf{3.35} & \textbf{2.53} & 1.85 \\
Blending-GRPO & 3.26 & 2.44 & \textbf{1.82} \\
\midrule
\textbf{HybridEmo} & 3.33 & 2.50 & 1.87 \\
\bottomrule
\end{tabular*}
\par
\begin{tabular*}{\columnwidth}{@{\extracolsep{\fill}}lccc@{}}
\toprule
\multicolumn{4}{l}{\textit{(b) Emotion blending}} \\
\cmidrule(lr){1-4}
Variant & Inten. $\uparrow$ & Nat. $\uparrow$ & WER (\%) $\downarrow$ \\
\midrule
CosyVoice 3 & 3.47 & 3.06 & \textbf{0.30} \\
\midrule
SFT & 3.50 & 3.03 & 0.50 \\
Direct Hybrid-GRPO & 3.50 & 3.03 & 0.71 \\
Trajectory-GRPO & 3.53 & 3.05 & \textbf{0.30} \\
Blending-GRPO & 3.63 & 3.17 & 0.37 \\
\midrule
\textbf{HybridEmo} & \textbf{3.71} & \textbf{3.24} & 0.37 \\
w/o margin & 3.66 & 3.05 & 0.32 \\
\bottomrule
\end{tabular*}
}
\caption{Ablation results on MultiEmo-Test. Trajectory Corr. averages 1E--3E and trajectory Nat. averages 2E--3E. Inten./Nat. denote blending intensity/naturalness; w/o margin removes the weaker-target margin.}
\label{tab:training_analysis}
\end{table}

SFT alone gives limited gains, while Direct Hybrid-GRPO improves some emotion scores but raises WER and remains below HybridEmo, supporting the role of supervised initialization before RL. Specialized variants are task-dependent: Trajectory-GRPO slightly exceeds joint HybridEmo on trajectory correctness and naturalness, whereas Blending-GRPO improves blending-oriented perceptual scores over SFT. Joint HybridEmo achieves the highest blending intensity and naturalness while remaining competitive on trajectory control.

Removing the weaker-target margin reduces Intensity from 3.71 to 3.66 and Naturalness from 3.24 to 3.05, lowering their mean from 3.48 to 3.36. These results suggest that limiting single-anchor dominance improves both aspects of blending quality.

\subsection{Representation and Reward-Space Analysis}

\paragraph{Speech-token reconstruction.}
We examine how much emotion-related information is retained after discrete-token reconstruction. We encode and reconstruct the full RAVDESS speech corpus\footnote{\url{https://zenodo.org/records/1188976}}~\cite{livingstone2018ravdess}, which consists of studio-recorded emotional utterances, with the CosyVoice 3 speech tokenizer and decoder. For each utterance, we extract emotion2vec+ large~\cite{ma2023emotion2vec} embeddings from the original and reconstructed waveforms and compute their paired cosine similarity. The mean similarity of \textbf{0.9441} shows that reconstruction retains highly similar emotion representations in the emotion2vec+ large space.

\begin{figure}[!t]
    \centering
    \includegraphics[width=0.90\columnwidth]{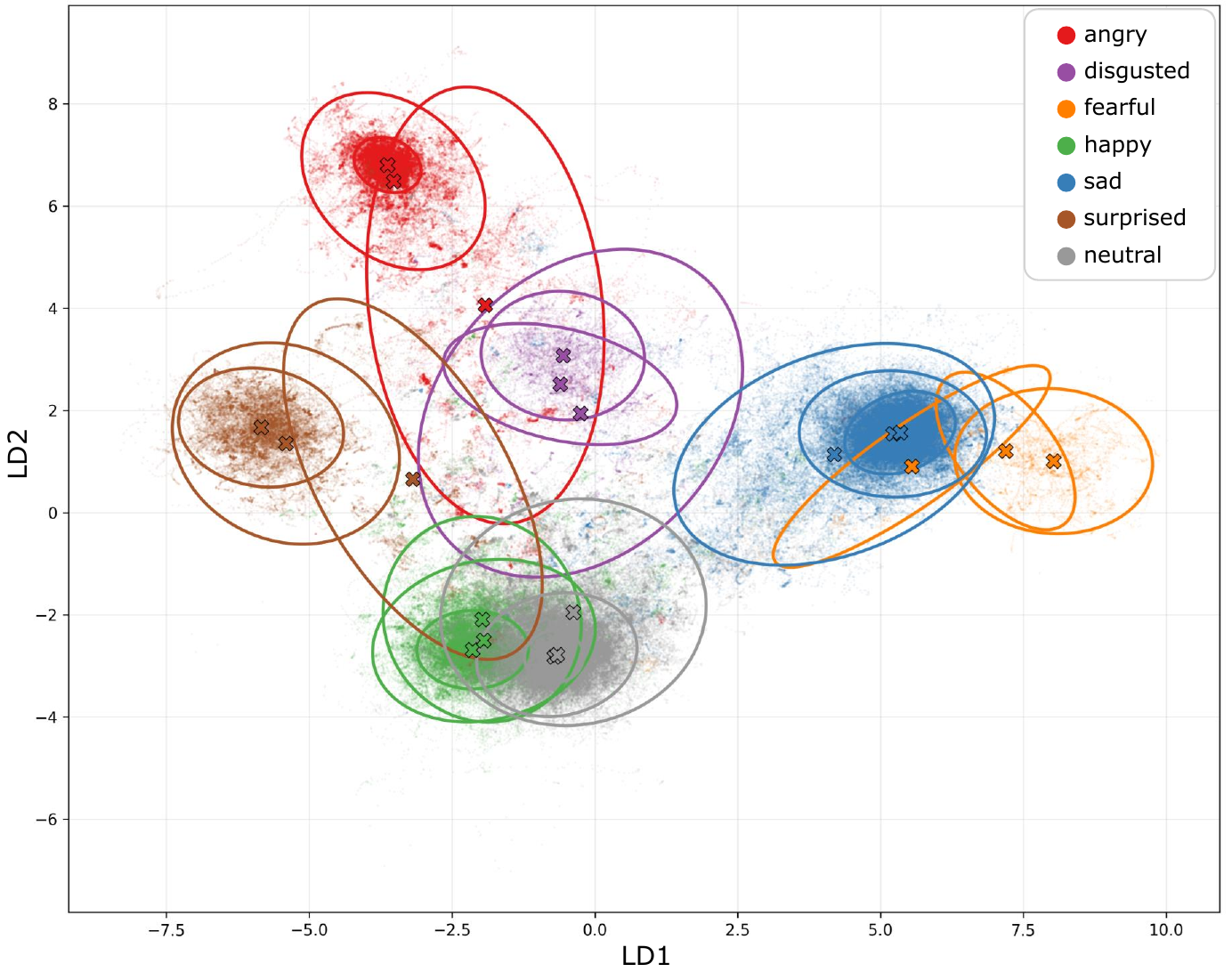}
    \caption{2D LDA projection of three-component GMM emotion anchors. Crosses mark component means, and ellipses show 95\% contours; scoring uses the full 6D space.}
    \label{fig:gmm_projection}
\end{figure}

\paragraph{Geometry of the GMM emotion anchors.}
Figure~\ref{fig:gmm_projection} visualizes the offline anchors fitted to filtered single-emotion utterances synthesized by CosyVoice 3. The projected distributions form distinct but partially overlapping regions, while multiple full-covariance components capture intra-emotion modes and anisotropic variation. These anchors provide continuous densities for mixture-density scoring. This geometry suggests that the anchors remain discriminative while allowing blended realizations to receive support from both target distributions.

\section{Conclusion}
\label{sec:conclusion}

In this paper, we formulate emotion control through emotion trajectories and blending. We propose HybridEmo, which combines supervised multi-emotion initialization with sample-aware hybrid-reward GRPO, routing task-matched rewards for trajectory and blending samples alongside shared ASR feedback. On our constructed MultiEmo-Test, HybridEmo significantly improves emotion correctness over CosyVoice 3 at every trajectory length and both blending-oriented perceptual scores, while matching Qwen3-TTS in blending intensity. It keeps WER below 2\% and UTMOS within 0.01 of CosyVoice 3 in both settings, without a noticeable degradation in speaker similarity. Human preferences further favor HybridEmo over CosyVoice 3 and EmoVoice-0.5B. The ablation study further demonstrates the complementary value of the two reward routes. The margin ablation also shows that the weaker-target margin, designed to limit single-anchor dominance, improves blending intensity and naturalness. Overall, these results show that matching reinforcement signals to task structure improves instruction-aligned perceptual outcomes across trajectory and blending settings, offering a practical direction for multi-emotion TTS.

\bibliography{references}

\end{document}